\documentclass[letterpaper, 10 pt, conference]{ieeeconf}  

\IEEEoverridecommandlockouts                              

\usepackage{graphics} 
\usepackage{amsmath,graphicx, booktabs, multirow, bbding}
\usepackage{subfigure}
\usepackage{mathrsfs}
\usepackage{color}
\allowdisplaybreaks[4]

\title{\LARGE \bf
Contrastive Learning for Mitochondria Segmentation
}

\author{Zhili Li, Xuejin Chen, Jie Zhao and Zhiwei Xiong
\thanks{Corresponding author: Xuejin Chen, xjchen99@ustc.edu.cn}%
\thanks{Zhili Li, Xuejin Chen, Jie Zhao and Zhiwei Xiong are with National Engineering Laboratory for Brain-Inspired Intelligence Technology and Application, University of Science and Technology of China and Institute of Artificial Intelligence, Hefei Comprehensive National Science Center, Hefei, China.}%
}

\begin{document}
\maketitle
\thispagestyle{empty}
\pagestyle{empty}

\begin{abstract}
Mitochondria segmentation in electron microscopy images is essential in neuroscience. However, due to the image degradation during the imaging process, the large variety of mitochondrial structures, as well as the presence of noise, artifacts and other sub-cellular structures, mitochondria segmentation is very challenging. In this paper, we propose a novel and effective contrastive learning framework to learn a better feature representation from hard examples to improve segmentation. 
Specifically, we adopt a point sampling strategy to pick out representative pixels from hard examples in the training phase. Based on these sampled pixels, we introduce a pixel-wise label-based contrastive loss which consists of a similarity loss term and a consistency loss term. The similarity term can increase the similarity of pixels from the same class and the separability of pixels from different classes in feature space, while the consistency term is able to enhance the sensitivity of the 3D model to changes in image content from frame to frame. We demonstrate the effectiveness of our method on MitoEM dataset as well as FIB-SEM dataset and show better or on par with state-of-the-art results. 
\end{abstract}

\section{INTRODUCTION}
\label{sec:intro}
Known as the powerhouses of cells, mitochondria play a crucial role in the regulation of cellular life and death for they carry out all types of important cellular functions by producing the overwhelming majority of cellular adenosine triphosphate (ATP). Many clinical studies have revealed the correlation between the biological function of mitochondria and their size and geometry~\cite{roychaudhuri2009amyloid,campello2010mitochondrial}. In the last few years, with the development of imaging technology, electron microscopy (EM) images have been widely used for the analysis of nanometer-level structures including mitochondria. Knott {\it et al.} \cite{knott2008serial} deliver a dataset of an adult rodent brain tissue using focused ion beam scanning electron microscopy (FIB-SEM) to see axons, dendrites and their synaptic contacts. More recently, Wei {\it et al.} \cite{wei2020mitoem} publish a large-scale EM dataset (MitoEM) for 3D mitochondria instance segmentation. MitoEM dataset is $3,600\times$ larger than FIB-SEM dataset \cite{knott2008serial} and consists of more mitochondria with sophisticated shapes and ultra-structures that are similar in appearance to mitochondria. Therefore, this dataset poses new challenges to the segmentation of mitochondria. 

Many earlier approaches combine traditional image processing and machine learning techniques for mitochondria segmentation. Jorstad and Fua \cite{10.1007/978-3-319-16220-1_26} propose an active surface-based method to refine the boundary surfaces by exploiting the thick and dark membranes of mitochondria. Lucchi {\it et al.} \cite{2013.259} utilize an approximate subgradient descent algorithm to minimize the hinge loss in the margin-sensitive frameworks. However, these methods rely on hand-crafted features to build the classifiers, limited in  efficiency and generalizability. 
Recently, deep learning-based approaches \cite{liu2018automatic, oztel2017mitochondria, Xiao-2018, cheng2017volume, casser2020fast, liu2020automatic} show promising results in mitochondria segmentation. Liu {\it et al.} \cite{liu2018automatic} adopt a modified Mask R-CNN for mitochondria instance segmentation on the FIB-SEM dataset \cite{knott2008serial}. Oztel {\it et al.} \cite{oztel2017mitochondria} propose a fully convolutional network (FCN) to segment mitochondria and then use various post-processing, such as 2D spurious detection filtering, boundary refinement and 3D filtering, to improve the segmentation. Xiao {\it et al.} \cite{Xiao-2018} propose an effective 3D residual fully convolutional network (FCN) to avert the vanishing gradient problem for mitochondria segmentation. 
Most of the methods listed above improve the segmentation by dedicatedly designing the network architecture, or adding various post-processing algorithms. However, these methods rarely pay extra attention to hard examples, such as noises, imaging artifacts and sub-cellular ultra-structures, which widely exist in EM images. As a result, these approaches can scarcely segment the hard examples correctly, preventing further improvement in mitochondria segmentation accuracy.

In the last few years, contrastive learning has been widely applied in computer vision tasks \cite{he2020momentum, chen2020a, khosla2020supervised, Zhao2020ContrastiveLF, 10.1007/978-3-030-59716-0_54}. Contrastive learning is an approach to formulate the task of finding similar and dissimilar things for a deep learning model. \cite{he2020momentum, chen2020a} adopt contrastive learning for unsupervised visual representation learning. For image classification, \cite{khosla2020supervised} extend the self-supervised batch contrastive approach to the fully-supervised setting, which effectively leverage label information. More recently, \cite{Zhao2020ContrastiveLF} propose to first pretrain the CNN feature extractor using a label-based contrastive loss for semantic segmentation task. \cite{10.1007/978-3-030-59716-0_54} adopt point-wise contrastive learning to improve boundary estimation for ultrasound image segmentation. Inspired by \cite{10.1007/978-3-030-59716-0_54}, we extend contrastive learning to solve hard examples in mitochondria segmentation, based on the intra-class similarity of mitochondria textures and the inter-class discrepancy of textures between mitochondria and hard examples. 

In this work, we propose a contrastive learning based framework for mitochondria segmentation task. In the training phase, we first sample representative points (i.e. pixels) from hard examples. Then we adopt a contrastive loss, which is computed on the feature space of these points, to facilitate model training. Our contrastive loss consists of two terms, similarity term and consistency term. The similarity loss term is designed to improve the feature similarity of the points from the same class and decrease that of different classes. The consistency loss term focuses on the feature similarity of points in adjacent frames to enhance model robustness of image content changes. We conduct many experiments on both MitoEM dataset and FIB-SEM dataset to verify the effectiveness of proposed method.
\begin{figure}[tbp]
\begin{minipage}[b]{1.0\linewidth}
  \centering
  \centerline{\includegraphics[scale=.26]{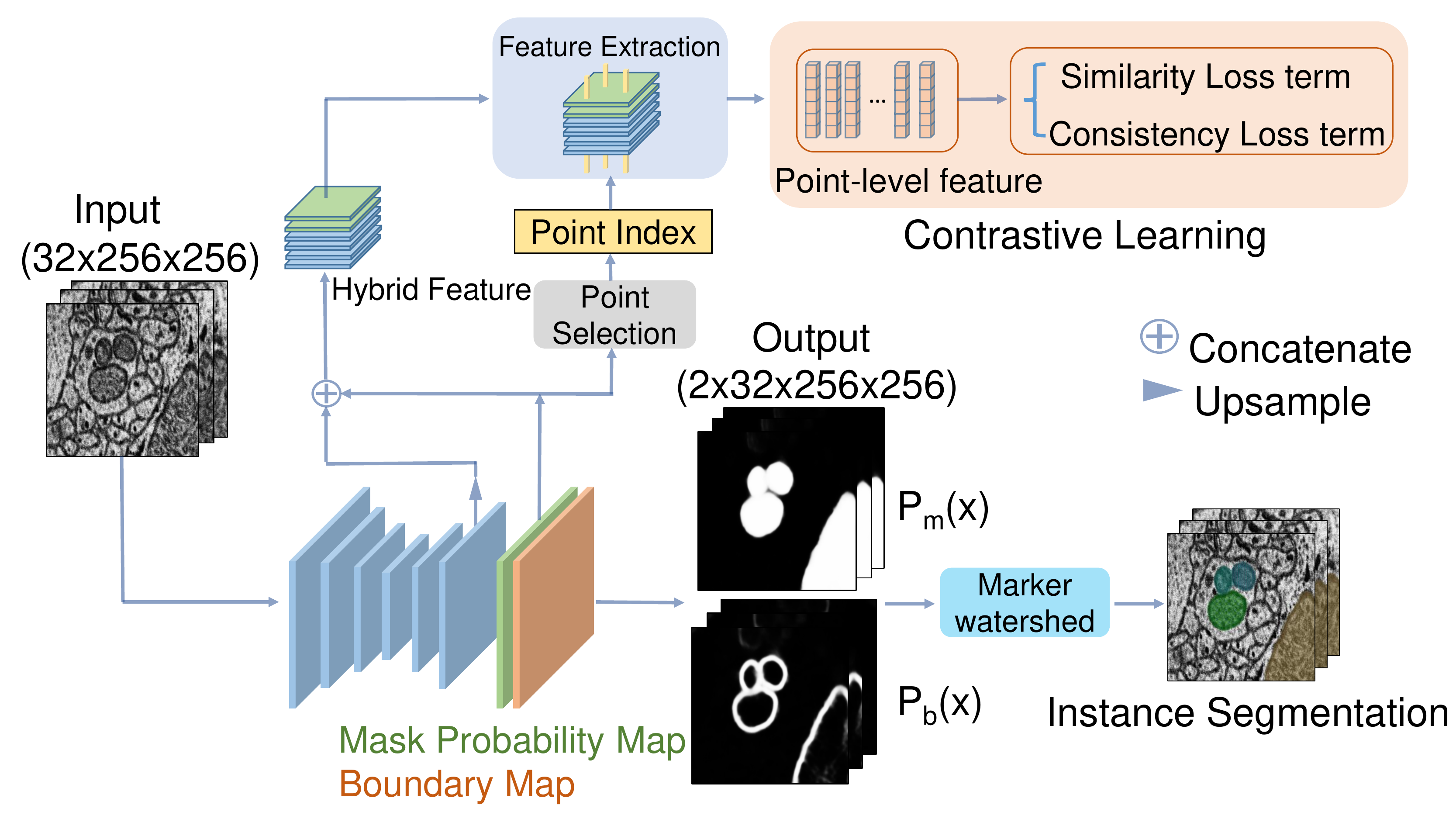}}
\end{minipage}
\caption{The pipeline of our proposed framework.}
\label{fig:overflow}
\end{figure}

\section{Our Method}
\label{sec:method}
Our method consists of three components: a backbone network in a symmetric encoder-decoder structure, a point selection block to sample representative points, and a contrastive learning head to compute the contrastive loss during the training stage, as shown in Fig. \ref{fig:overflow}. 

\subsection{Network Architecture}
We build a semantic segmentation backbone network based on the 3D U-Net \cite{3dunet}. 
Following \cite{DCAN}, we generate a mask probability map and a boundary map from the backbone segmentation network. At the end of our framework, a marker-controlled watershed post-processing algorithm \cite{marker-watershed} is applied to isolate each instance from the semantic segmentation map. 
In this paper, we denote our backbone network by U3D-BC following \cite{wei2020mitoem}.

Training the segmentation network greatly relies on the labeled data which is expected to be diverse and representative.
For each input image $X\in R^{W\times H}$, we have its dense mask label $M\in \{0,1\}^{W\times H}$ as well as the boundary label $B\in \{0,1\}^{W\times H}$, which can be generated from the mask label, same as \cite{wei2020mitoem}. 
In previous methods, a common way to train the segmentation network is to minimize the overall cross-entropy loss defined as

\begin{small}
\begin{equation}
\begin{split}
L_{CE} = & \sum_{x \in X} -{\rm log}\ P_m(x)M(x)-{\rm log}\big(1-P_m(x)\big)(1-M(x))\\
& + \sum_{x \in X} -{\rm log}\ P_b(x)B(x)-{\rm log}\big(1-P_b(x)\big)(1-B(x)) \\
\end{split}
\label{equ: cross_entropy}
\end{equation}
\end{small}

where $x$ is a pixel of the image ${X}$, $P_m$ denotes the predicted probability map of mitochondria masks, and similarly $P_b$ is the predicted probability of boundaries.

The cross-entropy loss penalizes the predicted probability of all pixels without distinction. Moreover, there are usually far more easy examples than hard samples in EM images. Therefore, it's challenge for existing methods to learn good representations of hard examples, making the predictions of them inaccuracy. To solve this problem, we propose to first select representative points from both hard examples and easy examples, and then penalize the predictions of hard points additionally in a contrastive manner. 
In this way, hard examples are paid extra attention and consequently their segmentation results can be possibly aided.

\subsection{Point Selection Block}
\label{point selection}
The point selection block is designed to sample representative points during the training phase.
We sample two types of points from the input image. One is uncertain points, which are sampled from hard examples, and the other one is certain points, which have high prediction confidence in the mask probability map. 
Let $N$ denote the number of sampled points, then the specific process of point selection is as follows.
Based on the mask probability map, we randomly sample a representative point set $\mathcal{S}$, which consists of $\beta N$ uncertain points and $(1-\beta)N$ certain points. The uncertain points are sampled based on the prediction error, which is the absolute value of the difference between the predicted probability and the ground truth label. Top $\beta N$ points with the highest prediction error make up uncertain point set $\mathcal{S}_u$. Likewise, we choose the top $(1-\beta)N$ points with the highest  mask probability to constitute the certain point set $\mathcal{S}_c$. 
\subsection{Contrastive Learning}
\label{cl}
With $N$ representative points, we adopt contrastive learning to mine hard examples in the training phase. Contrastive learning is mostly utilized in feature space to facilitate model training. 
The point features are derived from the backbone network. In the backbone segmentation network, we extract a mask probability map $M$ and a feature map $F\in R^{\frac{W}{2}\times \frac{H}{2} \times C}$ from the last two layers. 
We use trilinear interpolation to upsample the feature map to size of $W\times H$ and concatenate it with the probability map $M$ into a hybrid feature map. The feature vectors of representative points in the hybrid feature map can be extracted according to their point index. 

With the feature vectors of $N$ representative points, we compute a contrastive loss to ``weight'' hard points. Our contrastive loss consists of two terms: similarity loss term and consistency loss term. The similarity term can improve the feature similarity of points from the same class, while reducing that of points from different classes. Thus the segmentation of hard examples can be possibly aided. Complementary to the similarity term, the consistency term is designed to enhance the feature similarity between points of the same class at the same position in adjacent frames, and contrastively decrease the similarity of those from different classes. As a result, the inter-frame continuity of segmentation predictions will be improved. Both of the two loss terms contribute to a more discriminative and robust model. 

The steps of similarity loss term computation are as follows. First, we split point set $\mathcal{S}$ into three sets, certain foreground set $\mathcal{S}_{cf}$, uncertain foreground set $\mathcal{S}_{uf}$ and background set $\mathcal{S}_{b}$ according to their predicted probability and class label. Every point labeled as foreground in the certain point set $\mathcal{S}_c$ will be clustered to $\mathcal{S}_{cf}$. Similar clustering rule is also applied to the uncertain point set $\mathcal{S}_u$: for each point in $\mathcal{S}_u$, if it belongs to the foreground then it will be assigned to $\mathcal{S}_{uf}$. The remaining points in $\mathcal{S}$ are clustered to $\mathcal{S}_{b}$.
Afterwards, the certain foreground set and the uncertain foreground set form the positive pair while the certain foreground set and the background set build the negative pair.
Let $\mathbf{p}_r$ represent the feature vector of point $r$ and $r_z$ be the normalized $z$ coordinate of point $r$. Then the similarity loss $L_{sim}$ is computed in the following formula:
\begin{align}
    & Sim(\mathbf{p},\mathbf{q}) = \frac{\mathbf{p}\cdot\mathbf{q}}{\|{\mathbf{p}}\| \cdot \|{\mathbf{q}}\|}  \label{sim}\\ 
    & w(r, s) = {\rm exp}(\alpha \cdot(r_z-s_z)^2) \\
    & pos(r)  = \frac{1}{|\mathcal{S}_{uf}|}\sum_{s \in \mathcal{S}_{uf}}w(r, s)\cdot {\rm exp}(Sim(\mathbf{p}_r, \mathbf{p}_s)) \\
    & neg(r) = \frac{1}{|\mathcal{S}_{b}|}\sum_{t \in \mathcal{S}_{b}}w(r, t)\cdot {\rm exp}(Sim(\mathbf{p}_r, \mathbf{p}_t)) \\
    & L_{sim}=\gamma - \frac{1}{|\mathcal{S}_{cf}|}\sum_{r \in \mathcal{S}_{cf}}{\rm log}[1+\frac{pos(r)}{neg(r)}]
\end{align} 
We adopt $Sim(\mathbf{u}, \mathbf{v})$ to compute the cosine similarity between two vectors $\mathbf{u}$ and $\mathbf{v}$. {\bf $|$·$|$} denotes the number of elements in a set. $w(r, s)$ is the similarity weight based on the distance along $z$ axis between point $r$ and point $s$. $\alpha$ is a constant term that adjusts the correlation of features between adjacent frames. $\gamma$ is a constant used to ensure that $L_{sim}$ is always greater than zero.

The consistency loss term is computed as follows. For each point $s$ with coordinates $(x,y,z)$ in the sampled point set $\mathcal{S}$, we sample another point $s'$ on $(x,y,z-1)$ from adjacent frame. If point $s$ has the same class label with point $s'$, they will be viewed as a positive pair and otherwise be regarded as a negative pair. We denote the positive pair set as $\mathcal{P}$, and the negative pair set as $\mathcal{N}$. The formulation of loss term $L_{con}$ can be written down as follows:
\begin{align}
    & pos=\frac{1}{|\mathcal{P}|}\sum_{(s, s') \in \mathcal{P}}{\rm exp}(Sim(\mathbf {p}_s, \mathbf {p}_s')) \\
    & neg=\frac{1}{|\mathcal{N}|}\sum_{(s, s') \in \mathcal{N}}{\rm exp}(Sim(\mathbf {p}_s, \mathbf {p}_s')) \\
    & L_{con} = {\rm log}(1+\frac{pos}{neg})
\end{align}
The overall loss function of our framework is:
\begin{equation}
    L_{total} = L_{CE} + \lambda_1 L_{sim} + \lambda_2 L_{con}
\end{equation}
$L_{CE}$ denotes the cross entropy loss on the output mask probability map and boundary map. $\lambda_1$ and $\lambda_2$ are two constant terms that control the weight of $L_{sim}$ and $L_{con}$, separately.

\section{Experiments and Results}
\label{sec:exp}
In this section, we first introduce some details related to the experimental datasets and implementation. Then the results of our method and the quantitative and qualitative comparisons with other approaches are presented.
\subsection{Datasets}
We evaluate our method on two datasets (MitoEM dataset and FIB-SEM dataset). The related details are depicted as follows:

{\bf MitoEM dataset \cite{wei2020mitoem}}: This dataset consists of two $30 \ \mu m^3$ volumes, $1000\times4096\times4096$ in voxels at $30\times8\times8 \ nm^3$ resolution. The image volumes are acquired from a rat (MitoEM-R) and a human (MitoEM-H) tissue, respectively. For each image stack, only the labels of the first 500 slices have been published, and the labels of the remaining 500 slices are unreleased. In our experiments, we use the first 400 slices of published data for training and evaluate the segmentation performance on the last 100 slices.

{\bf FIB-SEM dataset \cite{knott2008serial}}: This dataset is obtained from mouse hippocampus and composed by a training volume and a testing volume. Each volume with a resolution of $5\times5\times5 \ nm^3/voxel$ consists of 165 slices and the size of each slice is $1024\times 768$.

\subsection{Implementation Details}
We implement the proposed method using the Pytorch open-source deep learning library \cite{pytorch}. In our experiments, the input data size for training is $32\times256\times256$ for both MitoEM and FIB-SEM dataset. The network is trained using the stochastic gradient descent \cite{sgd}. We follow the same data augmentation and learning schedule setting as \cite{superhuman}. We set the hyper-parameters in this work as follows: $N$=1024, $\beta$=0.75, $\alpha$=-4.0, $\lambda_1$=$\lambda_2$=0.2, $\gamma$=$\rm log(1+e^2)$. 

\begin{table}[btp]
\caption{Segmentation Results on MitoEM dataset.}
\label{tab:result}
\begin{center}
\setlength{\tabcolsep}{2.57mm}{
\begin{tabular}{cccccc} \toprule
{\multirow{2}*{Dataset}} & {\multirow{2}*{Method}} & \multicolumn{4}{c}{AP-75} \\
\cline{3-6}
{}&{}&  Small &  Medium & Large &  All \\ \hline
\multirow{2}* {MitoEM-R} & U3D-BC  & 0.139 & 0.724 & 0.895 & 0.844 \\
{} & Our method  & \bf 0.203 &  \bf 0.743 & \bf 0.913 & \bf 0.870 \\ \hline
\multirow{2}* {MitoEM-H} & U3D-BC  & 0.232 & 0.747 & 0.825 & 0.773 \\
{} & Our method  & \bf 0.296 & \bf 0.778 & \bf 0.830 & \bf 0.787 \\
\bottomrule
\end{tabular}}
\end{center}
\end{table}

\renewcommand{\arraystretch}{1.25}
\begin{table}[tbp]
\caption{Effect of contrastive loss terms on MitoEM dataset.}
\label{tab:ablation}
\begin{center}
\setlength{\tabcolsep}{1.4mm}
{
\begin{tabular}{cccccccc} \toprule
{\multirow{2}*{Dataset}} &  {\multirow{2}*{$L_{CE}$}} & {\multirow{2}*{$L_{sim}$}} & {\multirow{2}*{$L_{con}$}} & \multicolumn{4}{c}{AP-75} \\
\cline{5-8}
{} & {} & {} & {} & {Small} & {Medium} & {Large} & {All} \\ \hline
{\multirow{4}*{MitoEM-R}} & \scriptsize \Checkmark & \scriptsize\XSolidBrush & \scriptsize\XSolidBrush & 0.139 & 0.724 & 0.895 & 0.844 \\
{}  & \scriptsize \Checkmark & \scriptsize \XSolidBrush & \scriptsize \Checkmark & 0.166 & 0.679 & 0.909 & 0.858 \\
{}  & \scriptsize \Checkmark & \scriptsize\Checkmark & \scriptsize\XSolidBrush & 0.185 & \bf 0.747 & 0.909 & 0.866 \\
{}  & \scriptsize \Checkmark & \scriptsize\Checkmark & \scriptsize\Checkmark & \bf 0.203 & 0.743 & \bf 0.913 & \bf 0.870 \\ \hline
{\multirow{4}*{MitoEM-H}} & \scriptsize \Checkmark & \scriptsize\XSolidBrush & \scriptsize\XSolidBrush & 0.232 & 0.747 & 0.825 & 0.773 \\
{}  & \scriptsize \Checkmark & \scriptsize\XSolidBrush & \scriptsize\Checkmark & 0.216 & 0.755 & 0.829 & 0.778 \\
{}  & \scriptsize \Checkmark & \scriptsize\Checkmark & \scriptsize\XSolidBrush & 0.292 & 0.777 & 0.825 & 0.781 \\
{}  & \scriptsize \Checkmark & \scriptsize\Checkmark & \scriptsize\Checkmark & \bf 0.296 & \bf 0.778 & \bf 0.830 & \bf 0.787 \\
\bottomrule
\end{tabular}}
\end{center}
\end{table}
\renewcommand{\arraystretch}{1}

\subsection{Evaluation Metrics and Results}
For segmentation evaluation, AP-75 and Jaccard index are two main metrics to measure the results. AP-75 evaluates the segmentation performance in instance level. In AP-75 evaluation metric, a predicted instance is considered as a TP only if the voxel-wise overlap between the prediction and corresponding ground truth reaches at least 75\%. Jaccard index is a common segmentation criteria to measure the accuracy in pixel level, and its formula is as follows: 
\begin{equation}
    {Jaccard \ index} = \frac{TP}{(TP+FP+FN)}
\end{equation}

Following \cite{wei2020mitoem}, we report the AP-75 results for small, medium, large and all instances separately to evaluate the segmentation results on MitoEM dataset. Table \ref{tab:result} shows the performance comparison of U3D-BC and our method on MitoEM-H and MitoEM-R validation datasets. It's obvious to see that our method brings continuous improvement over baseline on all indicators for both two datasets, which demonstrates the effectiveness of our approach. We also conduct ablation study about the effect of two contrastive loss terms on MitoEM dataset as shown in Table \ref{tab:ablation}. When utilizing similarity loss term ($L_{sim}$) or consistency loss term ($L_{con}$), our method can always achieve better results over the baseline. After combining them together, the segmentation performance can be further improved. This suggests that our two contrastive loss terms are essential and indispensable for training a good segmentor. 

\begin{figure}[tbp]
\begin{minipage}[b]{0.23\linewidth}
  \centering
  \centerline{\includegraphics[width=2.0cm]{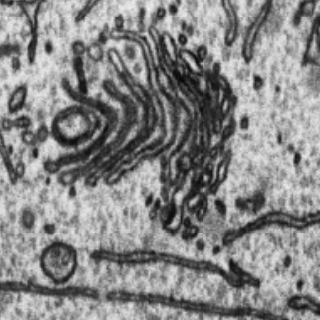}} 
  \vspace{.1cm}
  \centerline{\includegraphics[width=2.0cm]{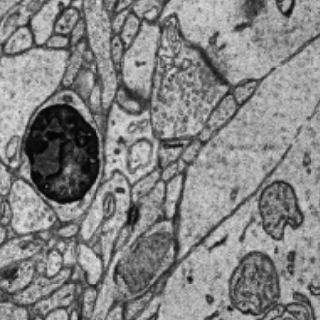}}
  \centerline{\small (a) Raw image}\medskip
\end{minipage}
\begin{minipage}[b]{0.23\linewidth}
  \centering
  \centerline{\includegraphics[width=2.0cm]{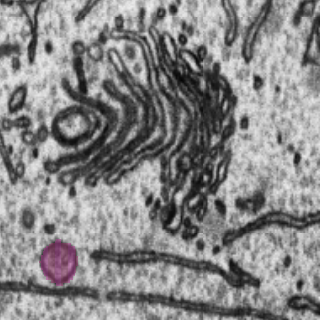}}
  \vspace{.1cm}
  \centerline{\includegraphics[width=2.0cm]{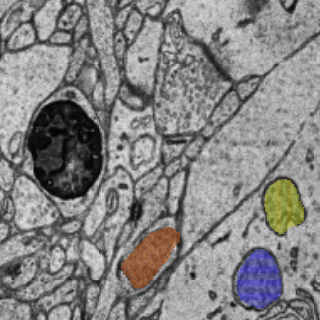}} 
  \centerline{\small (b) GT}\medskip
\end{minipage}
\begin{minipage}[b]{0.23\linewidth}
  \centering
  \centerline{\includegraphics[width=2.0cm]{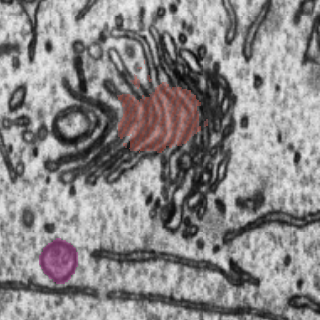}}
  \vspace{.1cm}
  \centerline{\includegraphics[width=2.0cm]{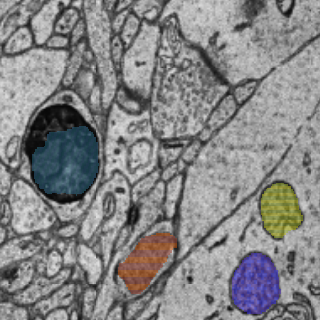}}
  \centerline{\small (c) U3D-BC}\medskip
\end{minipage}
\begin{minipage}[b]{0.23\linewidth}
  \centering
  \centerline{\includegraphics[width=2.0cm]{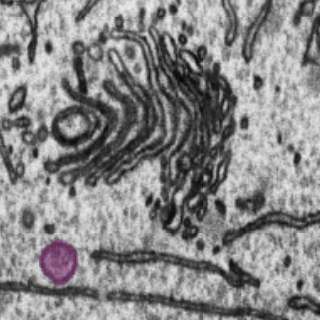}}
  \vspace{.1cm}
  \centerline{\includegraphics[width=2.0cm]{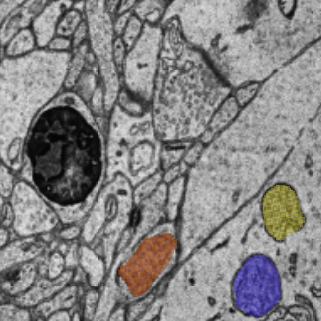}}
  \centerline{\small (d) Ours}\medskip
\end{minipage}
\caption{Visualization results on MitoEM dataset. From left to right: raw images, ground truth, results of U3D-BC, and results of our method.}
\label{fig:res}
\end{figure}

\begin{figure}[tbp]
\begin{minipage}[b]{0.48\linewidth}
  \centering
  \centerline{\includegraphics[width=4.2cm]{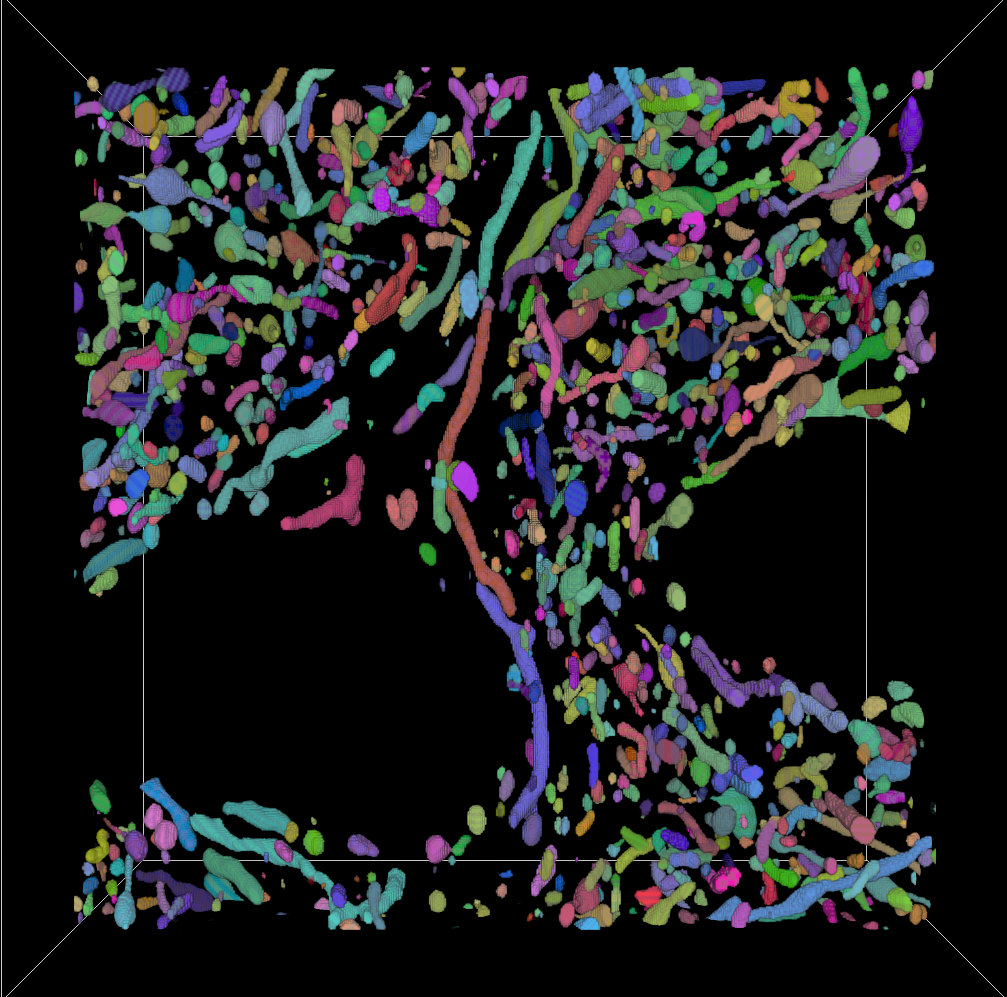}} 
  \centerline{(a) MitoEM-R}\medskip
\end{minipage}
\begin{minipage}[b]{0.48\linewidth}
  \centering
  \centerline{\includegraphics[width=4.2cm]{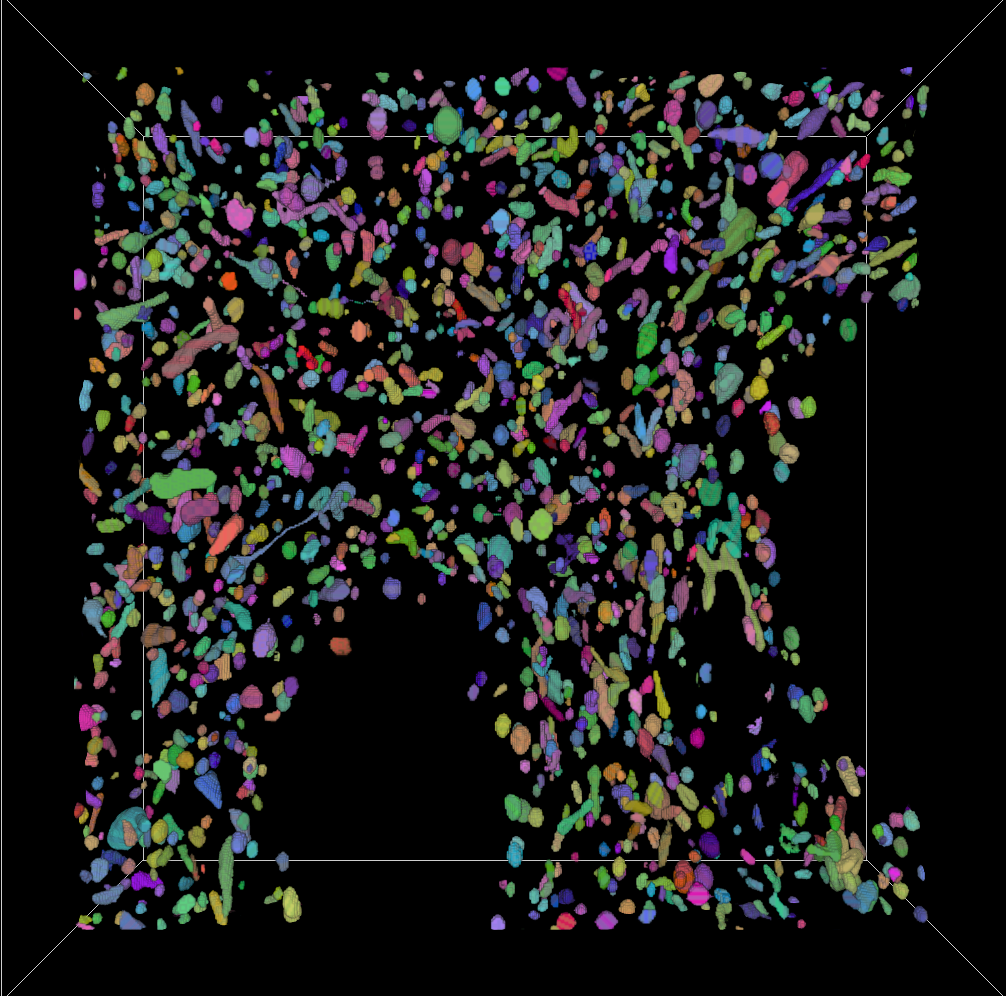}}
  \centerline{(b) MitoEM-H}\medskip
\end{minipage}
\caption{Instance segmentation results on MitoEM-R and MitoEM-H.}
\label{fig:instance_res}
\end{figure}

Fig. {\ref{fig:res}} illustrates the qualitative comparison of segmentation results between the baseline and our method on MitoEM dataset. Our method can correctly distinguish mitochondria-like organelles from real mitochondria. This indicates that our method can solve hard examples better. To clearly show the 3D mitochondria structure, we import our segmentation results into VAST \cite{vast} for visualization, as shown in Fig. \ref{fig:instance_res}.

\begin{figure*}[htbp]
\centering

\subfigure[Raw image]{
    \begin{minipage}[]{0.18\linewidth}
    \centering
    \includegraphics[width=1.32in]{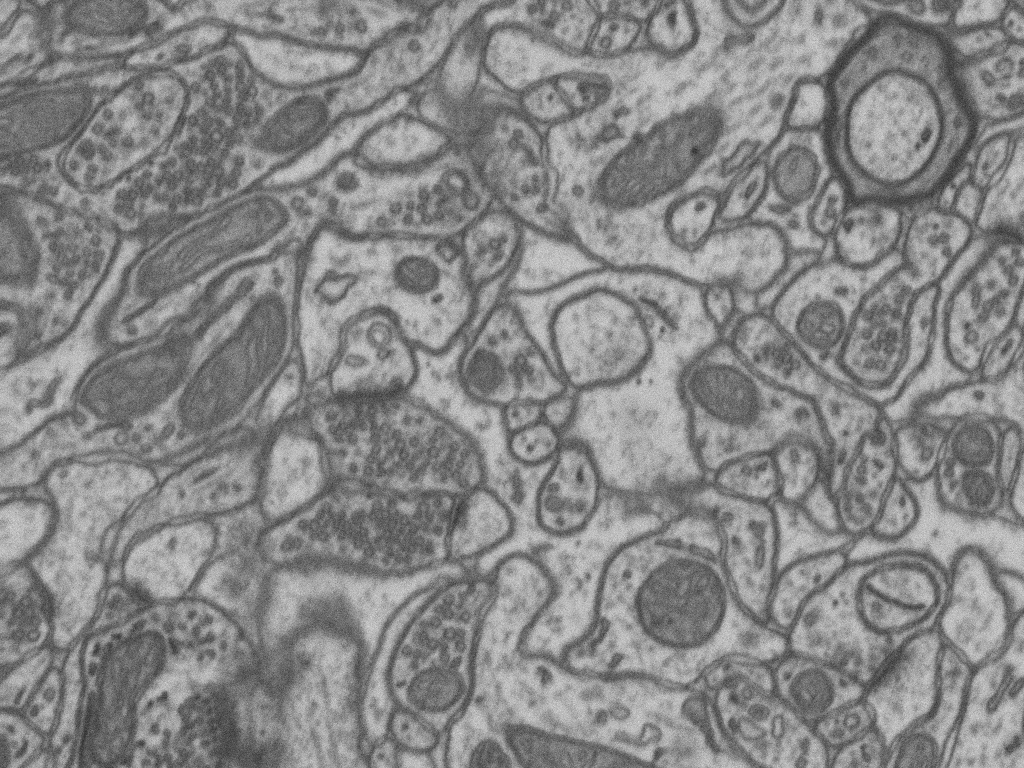} \\
    \vspace{0.2cm}
    \includegraphics[width=1.32in]{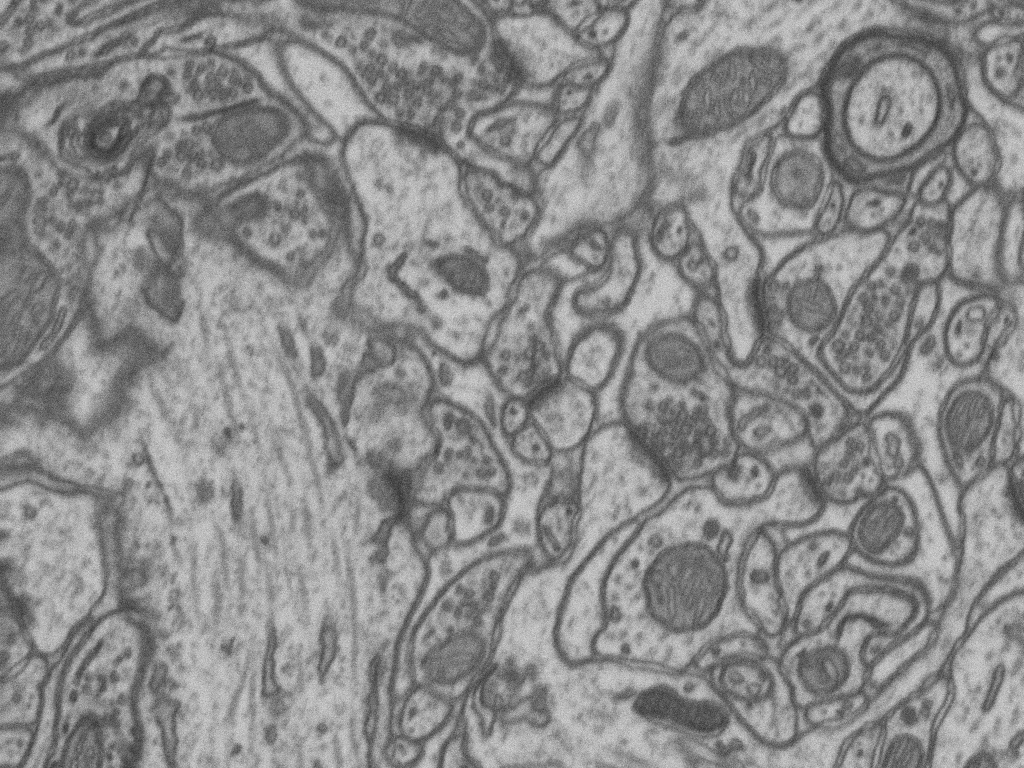} \\
    \vspace{0.2cm}
    \includegraphics[width=1.32in]{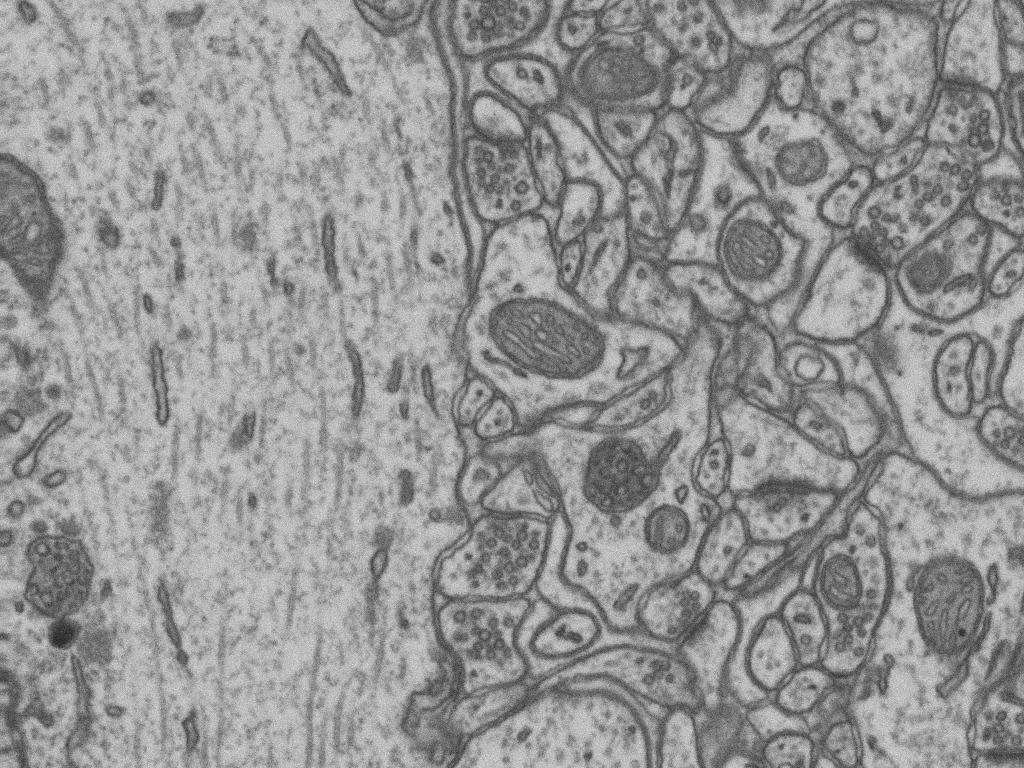} \\
    \vspace{0.2cm}
    \end{minipage}
}
\subfigure[Oztel {\it et al.} 2017 \cite{oztel2017mitochondria}]{
    \begin{minipage}[]{0.18\linewidth}
    \centering
    \includegraphics[width=1.32in]{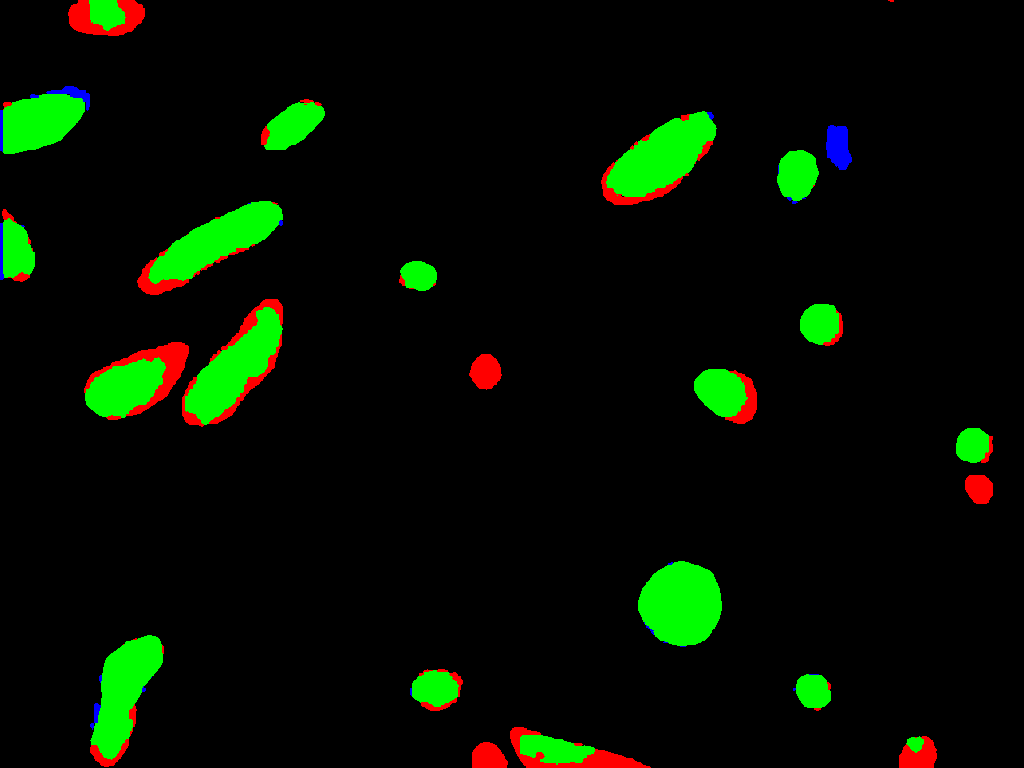} \\
    \vspace{0.2cm}
    \includegraphics[width=1.32in]{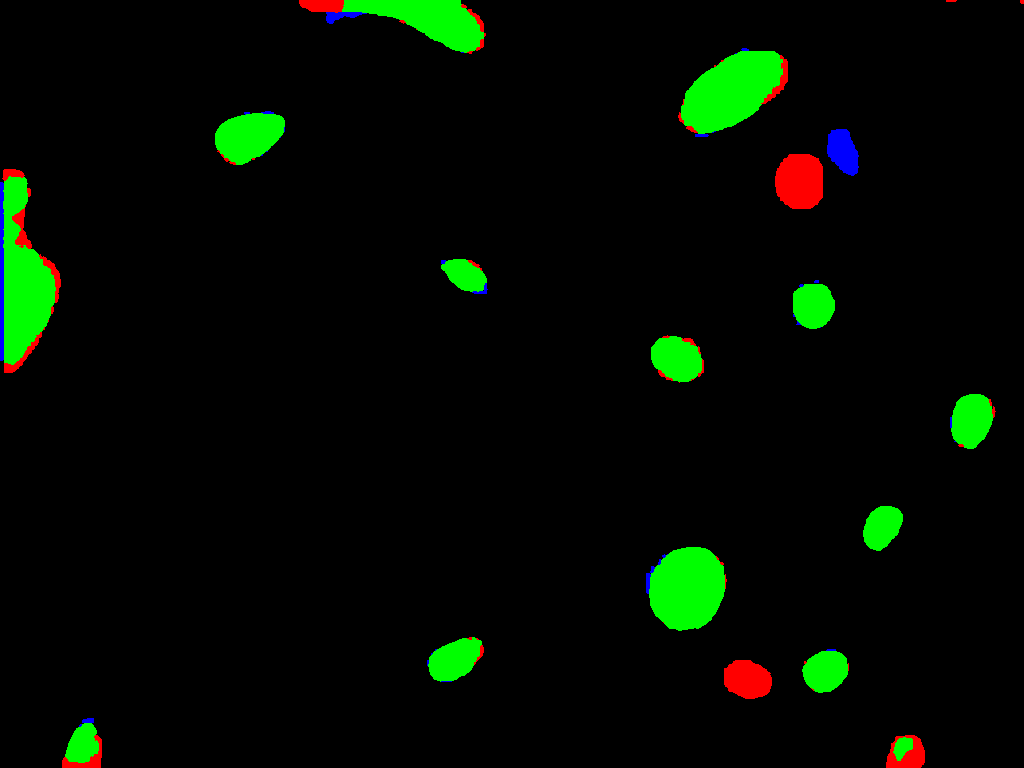} \\
    \vspace{0.2cm}
    \includegraphics[width=1.32in]{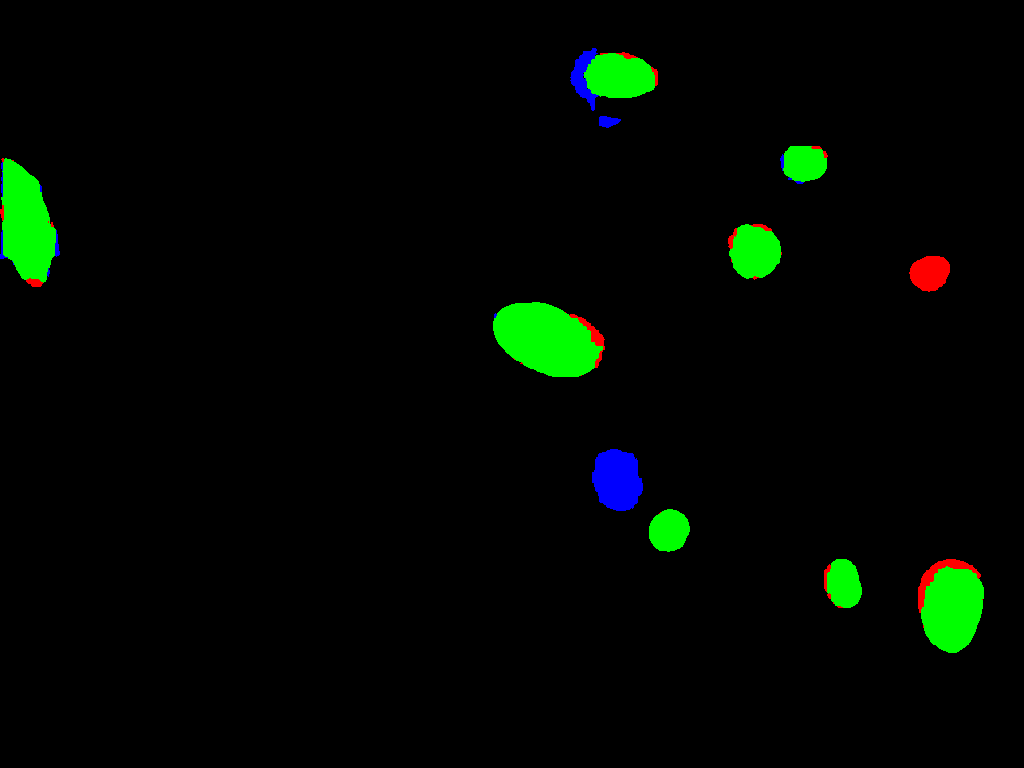} \\
    \vspace{0.2cm}
    \end{minipage}
}
\subfigure[Xiao {\it et al.} 2018 \cite{Xiao-2018}]{
    \begin{minipage}[]{0.18\linewidth}
    \centering
    \includegraphics[width=1.32in]{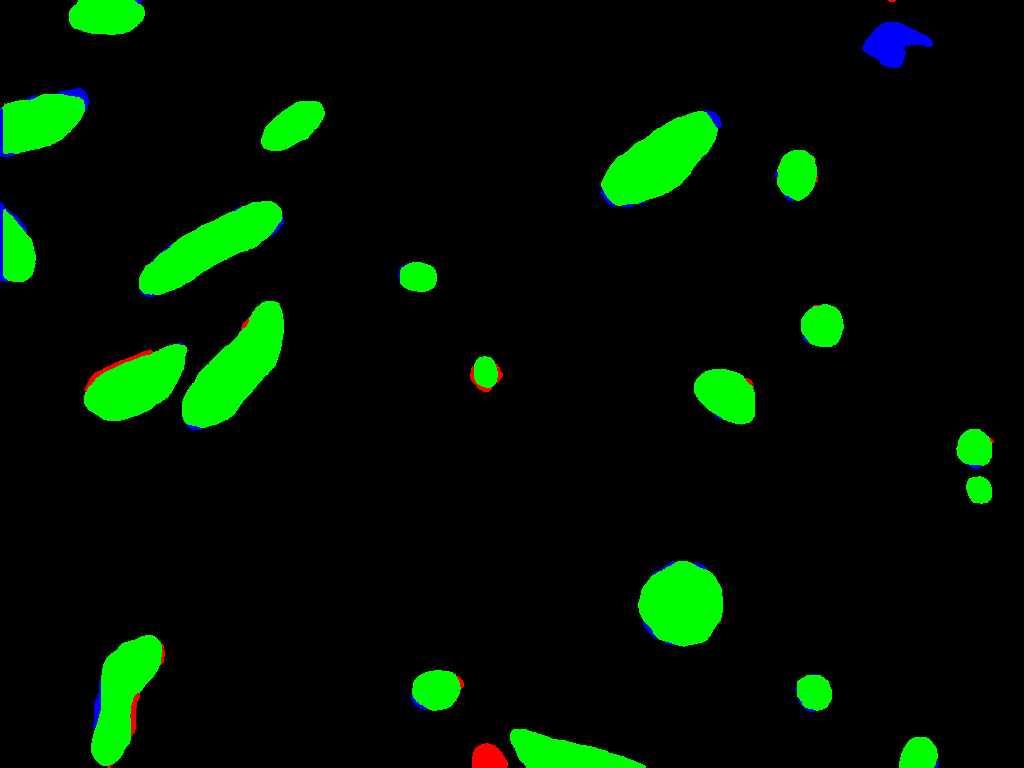} \\
    \vspace{0.2cm}
    \includegraphics[width=1.32in]{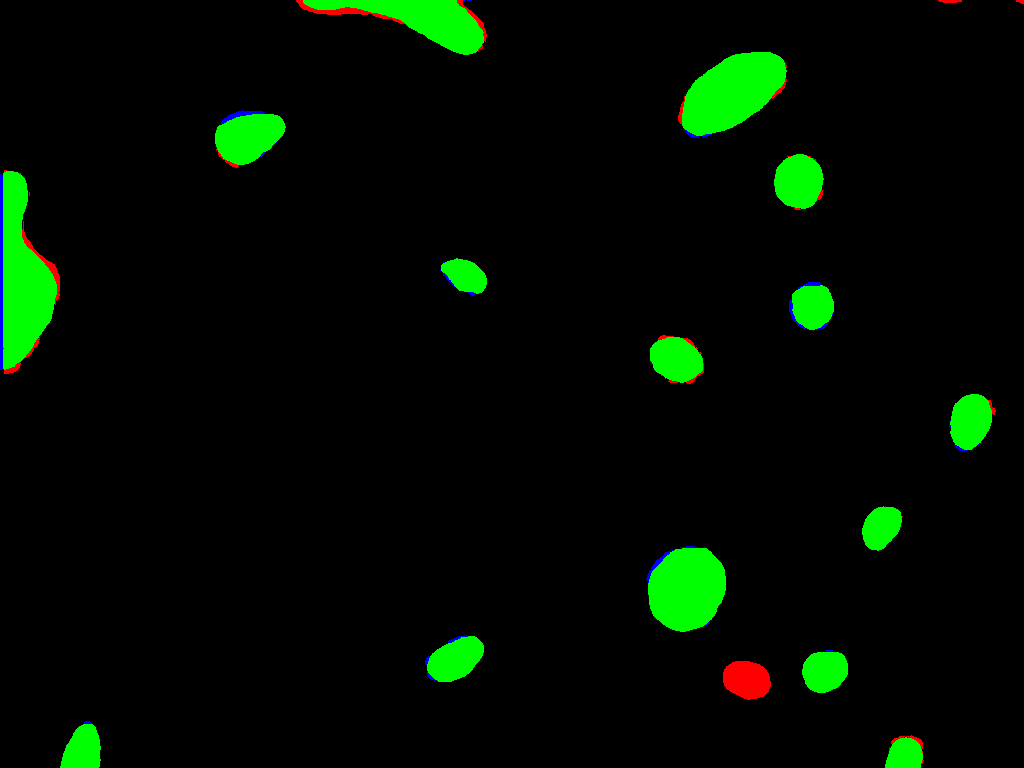} \\
    \vspace{0.2cm}
    \includegraphics[width=1.32in]{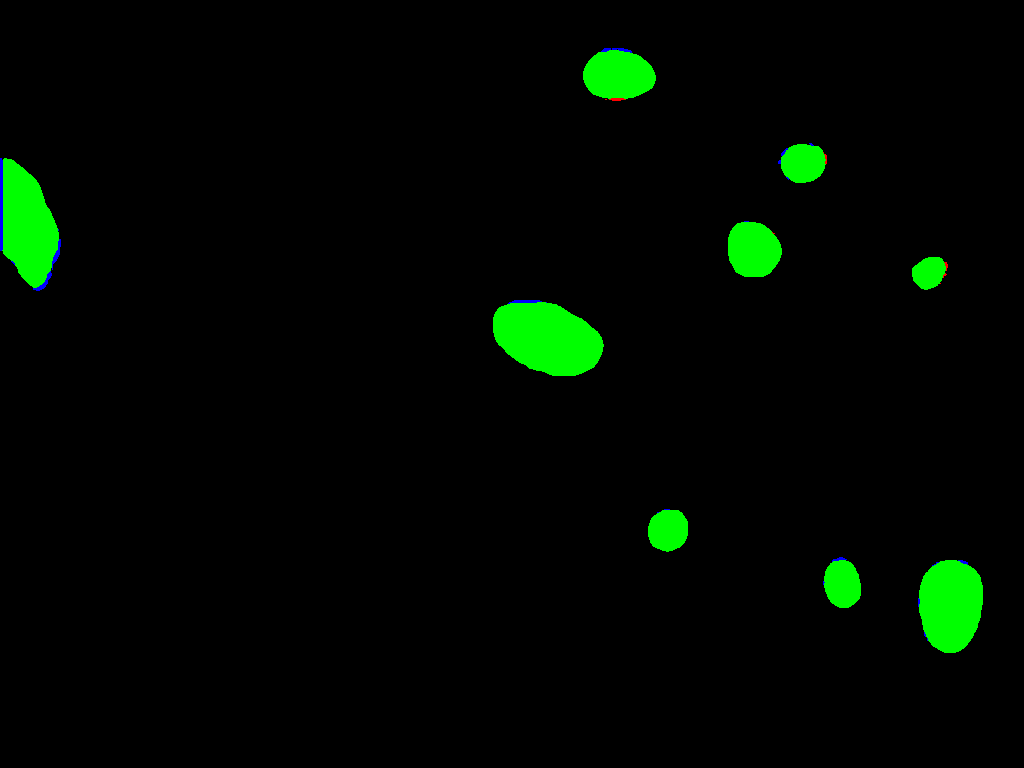} \\
    \vspace{0.2cm}
    \end{minipage}
}
\subfigure[Casser {\it et al.} 2020 \cite{casser2020fast}]{
    \begin{minipage}[]{0.18\linewidth}
    \centering
    \includegraphics[width=1.32in]{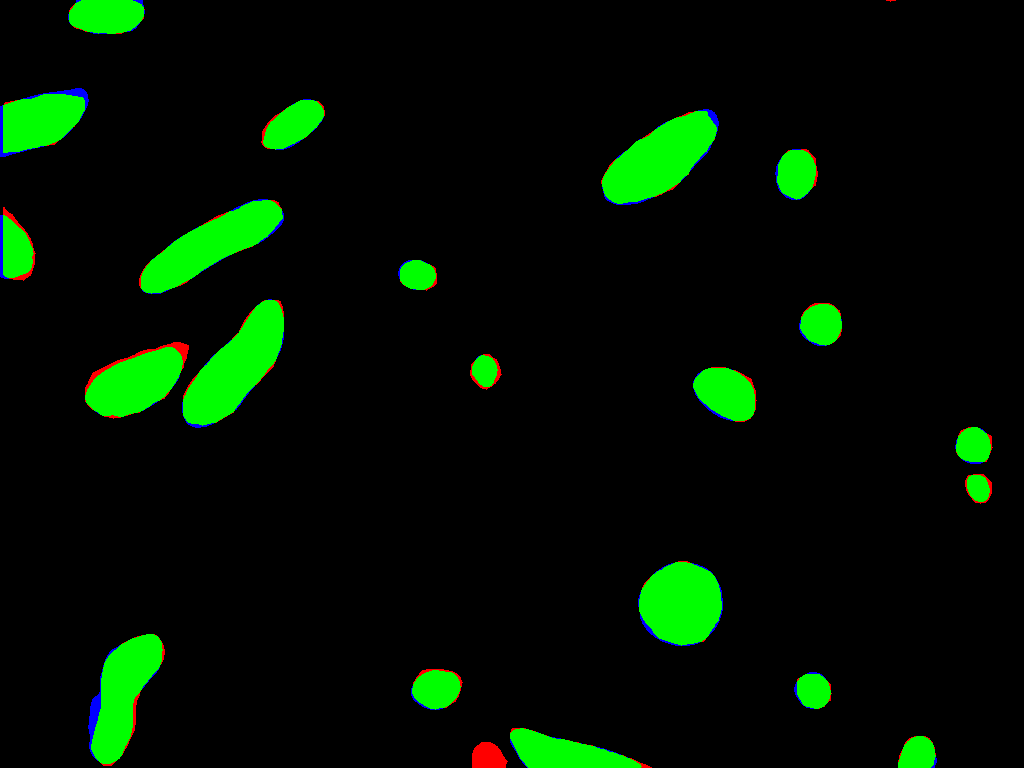} \\
    \vspace{0.2cm}
    \includegraphics[width=1.32in]{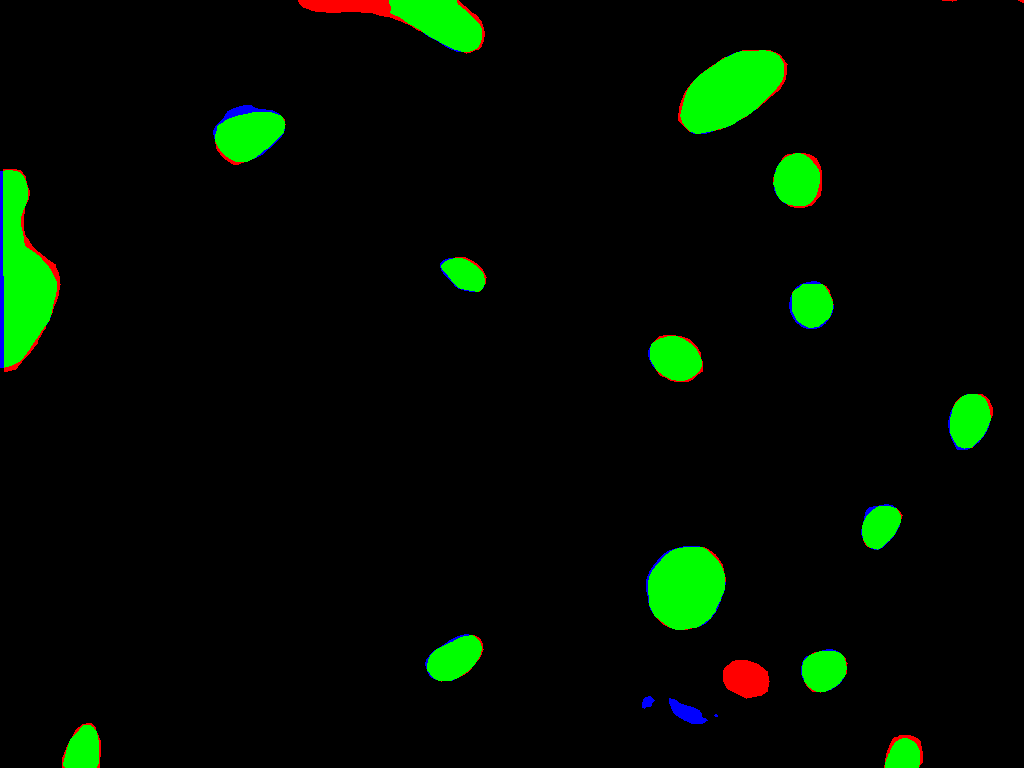} \\
    \vspace{0.2cm}
    \includegraphics[width=1.32in]{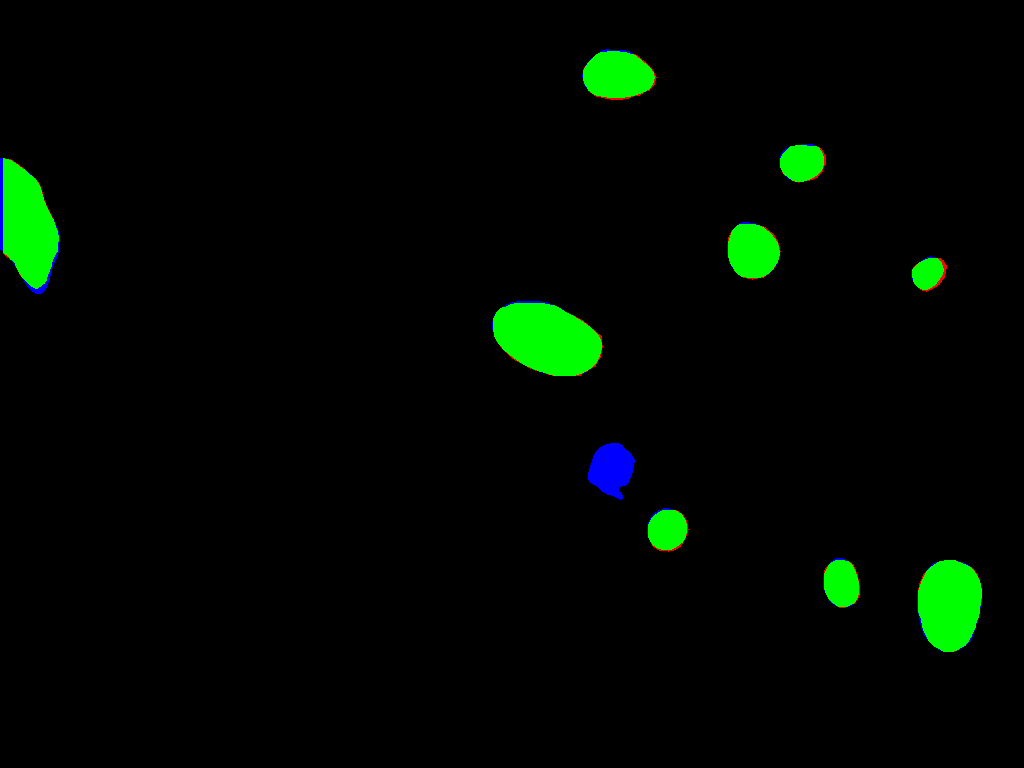} \\
    \vspace{0.2cm}
    \end{minipage}
}
\subfigure[Ours]{
    \begin{minipage}[]{0.18\linewidth}
    \centering
    \includegraphics[width=1.32in]{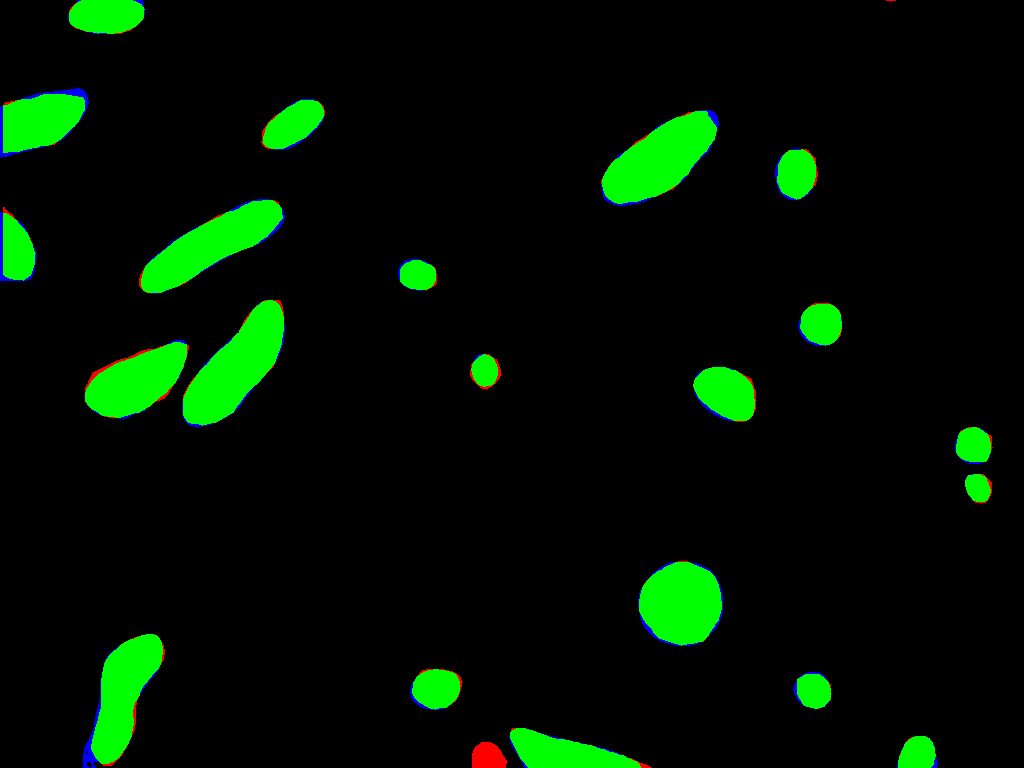} \\
    \vspace{0.2cm}
    \includegraphics[width=1.32in]{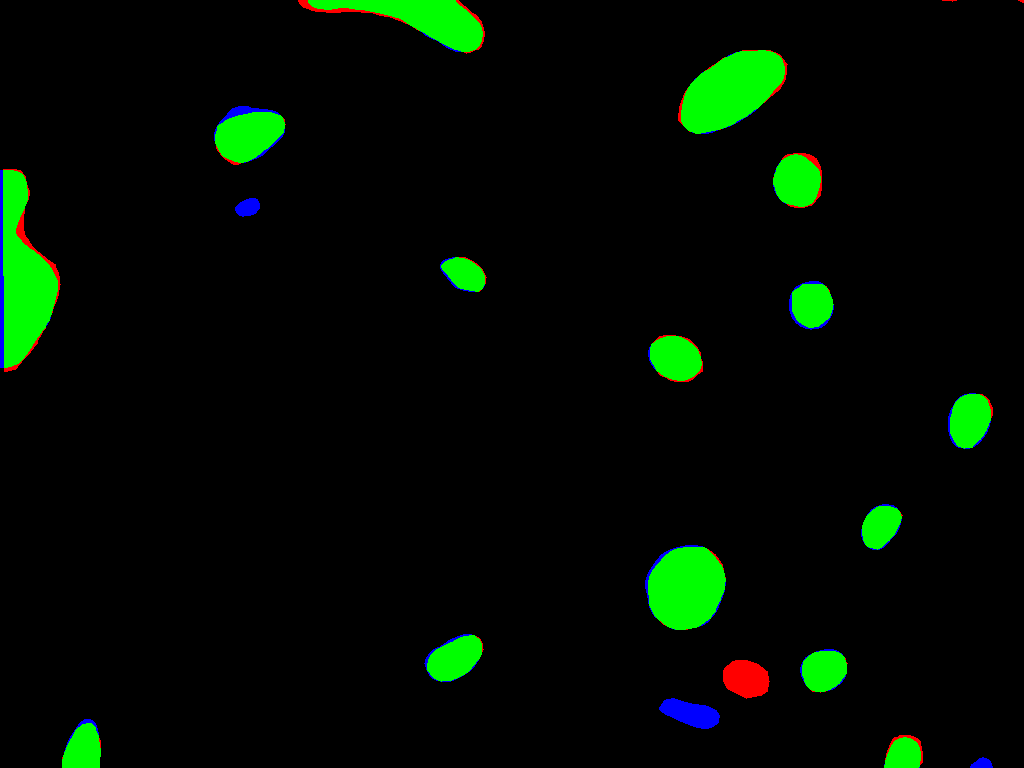} \\
    \vspace{0.2cm}
    \includegraphics[width=1.32in]{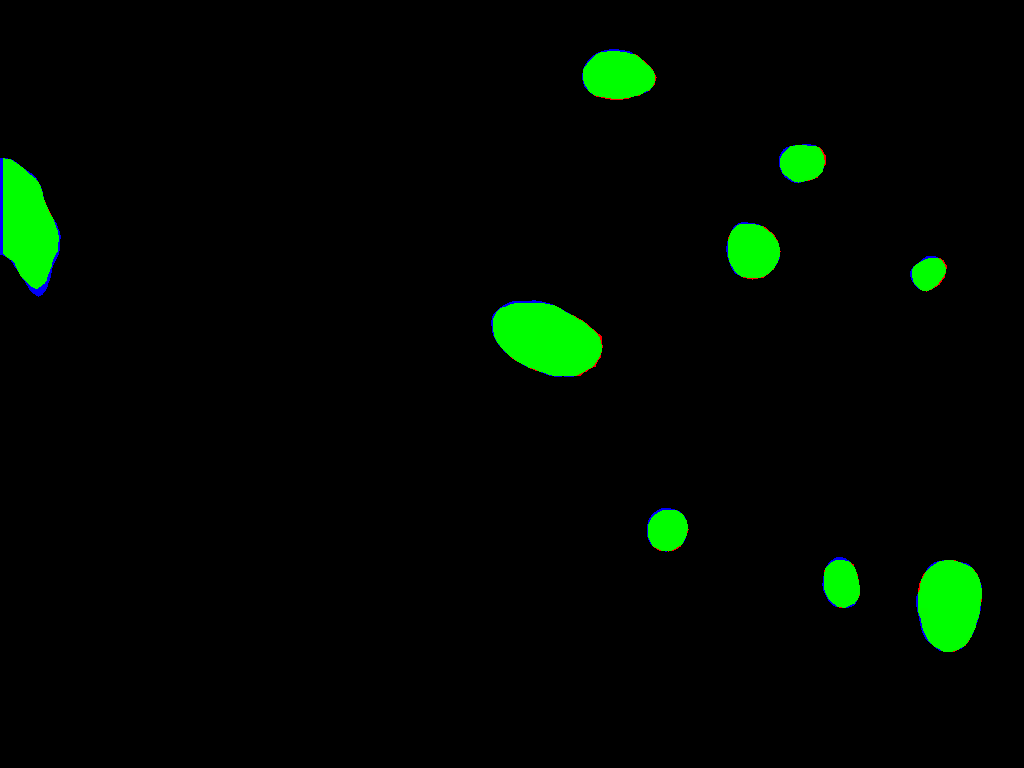} \\
    \vspace{0.2cm}
    \end{minipage}
}
\centering
\caption{The qualitative comparisons between different methods on FIB-SEM dataset. Green pixels denote true positives (TP), red pixels denote false negatives (FN), blue pixels denote false positives (FP), and black pixels denote true negatives (TN).}
\vspace{-0.2cm}
\label{fig:fib}
\end{figure*}

On FIB-SEM dataset, we utilize Jaccard index (foreground IoU) to measure segmentation accuracy following \cite{oztel2017mitochondria, Xiao-2018, cheng2017volume, casser2020fast}. The quantitative comparisons with previous methods are presented in Table \ref{tab:fib}, where the top three values are masked in bold for distinction. It can be seen that the Jaccard index of our approach is higher than most previous algorithms, demonstrating that our method achieves state-of-the-art results. To explicitly compare the qualitative segmentation differences between our method and others, we displayed 3 examples in Fig. \ref{fig:fib}. Note that our approach segments most of the mitochondria and effectively reduces FPs and FNs, achieving better qualitative results than \cite{oztel2017mitochondria, casser2020fast} and comparable results to \cite{Xiao-2018}.

\begin{table}[tbp]
\caption{segmentation results on FIB-SEM dataset}
\label{tab:fib}
\begin{center}
{\begin{tabular}{cc} 
\toprule
{\bf Method} & {\bf Jaccard index} \\ \midrule
Lucchi {\it et al.} 2013 \cite{2013.259}&  0.867  \\
3D U-Net \cite{3dunet}& 0.874  \\
Oztel {\it et al.} 2017 \cite{oztel2017mitochondria} & {\bf 0.907} \\
Cheng {\it et al.} 2017 \cite{cheng2017volume} & 0.889  \\
Improved Mask R-CNN \cite{liu2018automatic} &  0.849 \\
Xiao {\it et al.} 2018 \cite{Xiao-2018} & {\bf 0.900} \\
Casser {\it et al.} 2020 \cite{casser2020fast} & 0.890 \\
Liu {\it et al.} 2020 \cite{liu2020automatic} &  0.864 \\
Ours & {\bf 0.895} \\
\bottomrule
\end{tabular}}
\end{center}
\end{table}

\section{Conclusion}
\label{sec:conclusion}
In electron microscopy images, there are many hard examples which limit the improvement of mitochondria segmentation accuracy. To address this issue, we proposed an effective contrastive learning-based framework in which we first pick out representative points, then build positive pair and negative pair based on their labels and prediction probability, and finally compute a contrastive loss to facilitate model training. Our contrastive loss can increase intra-class compactness, inter-class separability and feature consistency between adjacent frames, thereby enabling a better feature representation. The experimental results on both two EM datasets demonstrate the effectiveness of our method in mitochondria segmentation. In the future, we plan to explore the application of contrastive learning in semi-supervised mitochondria segmentation task. 


\section{ACKNOWLEDGMENT}
This work was supported in part by National Key R\&D Program of China under Grant 2020AAA0105702, in part by the National Natural Science Foundation of China (NSFC) under Grants 62076230, in part by the University Synergy Innovation Program of Anhui Province GXXT-2019-025.

\end{document}